%% file: main.tex
\documentclass[journal]{IEEEtran}
\IEEEoverridecommandlockouts

\usepackage{booktabs}       
\usepackage{multirow} 

\usepackage{bbm} 
\usepackage{subcaption}
\usepackage{hyperref}

\usepackage{xspace}
\newcommand{\cotmix}{\textrm{CoTMix}\xspace} 

\usepackage{amsmath,amssymb,amsfonts}
\usepackage{graphicx}
\usepackage{balance}
\usepackage{tabularx}
\usepackage{booktabs}
\usepackage{breqn}  
\usepackage{nicematrix}

\usepackage{color, colortbl}	
\definecolor{LightCyan}{rgb}{0.88,1,1} 
    
\usepackage{float} 

\title{Contrastive Domain Adaptation for Time-Series via Temporal Mixup}

\author{Emadeldeen Eldele, Mohamed Ragab, Zhenghua Chen, Min Wu, Chee-Keong Kwoh and Xiaoli Li

\thanks{This work was supported by the Agency of Science Technology and Research under its AME Programmatic (Grant No. A20H6b0151) and its Career Development Award (Grant No. C210112046).}
\thanks{Emadeldeen Eldele is with the School of Computer Science and Engineering, Nanyang Technological University, Singapore and the Centre for Frontier AI Research, Agency for Science, Technology and Research, Singapore (E-mail: emad0002@ntu.edu.sg).}
\thanks{Mohamed Ragab is with the Institute for Infocomm Research, Agency for Science, Technology and Research, Singapore, Centre for Frontier AI Research, Agency for Science, Technology and Research, Singapore (E-mail: mohamedr002@e.ntu.edu.sg).}
\thanks{Zhenghua Chen is with the Institute for Infocomm Research, Agency for Science, Technology and Research, Singapore and the Centre for Frontier AI Research, Agency for Science, Technology and Research, Singapore (E-mail: chen0832@e.ntu.edu.sg).}
\thanks{Min Wu is with the Institute for Infocomm Research, Agency for Science, Technology and Research, Singapore (E-mail: wumin@i2r.a-star.edu.sg).}
\thanks{Chee-Keong Kwoh is with the School of Computer Science and Engineering, Nanyang Technological University, Singapore (E-mail: asckkwoh@ntu.edu.sg).}
\thanks{Xiaoli Li is with the Institute for Infocomm Research, Agency for Science, Technology and Research, Singapore, Centre for Frontier AI Research, Agency for Science, Technology and Research, Singapore, and also with the School of Computer Science and Engineering at Nanyang Technological University, Singapore (E-mail: xlli@i2r.a-star.edu.sg).}
\thanks{Min Wu is the corresponding author.}
}

\begin{document}
\markboth{Accepted in the \textbf{IEEE Transactions on Artificial Intelligence}}{Accepted in the \textbf{IEEE Transactions on Artificial Intelligence}}

\maketitle

\begin{abstract}

Unsupervised Domain Adaptation (UDA) has emerged as a powerful solution for the domain shift problem via transferring the knowledge from a labeled source domain to a shifted unlabeled target domain. Despite the prevalence of UDA for visual applications, it remains relatively less explored for time-series applications. In this work, we propose a novel lightweight contrastive domain adaptation framework called CoTMix for time-series data. Unlike existing approaches that either use statistical distances or adversarial techniques, we leverage contrastive learning solely to mitigate the distribution shift across the different domains. Specifically, we propose a novel temporal mixup strategy to generate two intermediate augmented views for the source and target domains. Subsequently, we leverage contrastive learning to maximize the similarity between each domain and its corresponding augmented view. The generated views consider the temporal dynamics of time-series data during the adaptation process while inheriting the semantics among the two domains. Hence, we gradually push both domains towards a common intermediate space, mitigating the distribution shift across them. Extensive experiments conducted on five real-world time-series datasets show that our approach can significantly outperform all state-of-the-art UDA methods. The implementation code of \cotmix is
available at \href{https://github.com/emadeldeen24/CoTMix}{github.com/emadeldeen24/CoTMix}.

\end{abstract}

\begin{impact}
Unsupervised domain adaptation (UDA) aims to reduce the gap between two related but shifted domains. Current UDA methods for time-series data are based on adversarial or discrepancy approaches. These methods are complex in training and cannot efficiently address the large domain shift. Therefore, in this work, we propose a time-series UDA framework based purely on contrastive learning, which is simpler in implementation and training. To leverage contrastive learning to mitigate domain shift, we propose a temporal mixup strategy to generate augmentations that are robust to the domain shift and can move both domains towards an intermediate domain. We show the efficacy of our proposed framework against baselines and validate the impact of our proposed temporal mixup against other augmentations.
\end{impact}



\section{Introduction}

The advance in deep learning has shown a significant performance improvement in many time-series applications e.g., healthcare and manufacturing. Unfortunately, such performance can only persist under the assumption that training and testing data are drawn from the same distribution. In reality, training and testing data can substantially vary in their temporal characteristics and working conditions, causing the deep learning model to significantly underperform. This phenomenon is well-known as the domain shift problem. Unsupervised Domain Adaptation (UDA) aims to reduce the domain shift by adapting a model trained on a labeled source domain to a shifted unlabeled target domain. Despite the dense literature on UDA for visual applications \cite{da_survey,da_survey2,tl_survey}, it is still less explored for time-series data.

Existing works in time-series UDA follow two mainstreams to adapt the source and target domains. One paradigm leverages a statistical distance such as maximum mean discrepancy (MMD) to minimize the discrepancy between source and target domains \cite{mmd}.  The other paradigm utilizes an adversarial scheme by training a domain discriminator to mitigate the domain shift \cite{codats,ts_da_attn,adast}. Despite the acclaimed performance of these approaches, they still suffer the following limitations.  First, they ignore the temporal dependencies in time-series data while matching the source and target distributions, leading to sub-optimal  adaptation performance. Second, most of the existing discrepancy-based approaches depend on reducing a distance measure, which may struggle to align distributions with large domain shifts \cite{dskn}. Third, the adversarial-based approaches are usually complex to train and rely on minimax optimization, which is hard to converge to a satisfactory local optimum \cite{adv_problem}. Last, both paradigms attempt to directly adapt the target domain distribution towards the source domain using the source domain knowledge, which can be less effective when aligning distant domains \cite{concensus}.

Meanwhile, contrastive learning has shown great success and proficiency in time-series representation learning tasks \cite{ts_tcc,ts2vec}. One of the key factors to this success is the careful design of augmentations \cite{simclr}. Existing augmentation techniques, e.g., adding noise, permutation, and time/frequency shift have shown competent performance in time-series representation learning tasks. However, a proper augmentation that can consider the temporal dependencies in time-series data while being robust to the distribution shift is yet to exist.

In this work, we propose a novel framework (\cotmix) that exploits contrastive learning solely to mitigate the domain shift in time-series data. The key motivation behind using contrastive loss solely is to provide a simple yet effective framework for unsupervised domain adaptation, that can be more powerful than discrepancy-based approaches while being less complex than adversarial-based approaches. Contrastive loss provides a powerful tool for learning a common feature representation between the source and target domains provided having suitable augmentations. In addition, the contrastive loss is particularly well-suited for unsupervised domain adaptation, as it does not require labeled data from the target domain, and its supervised version enables us to utilize the available labels in the source domain. However, since traditional augmentations are not well-suited to address the domain shift, we propose a novel cross-domain temporal mixup strategy to address this challenge. In particular, the temporal mixup strategy generates two new intermediate domains namely the source-dominant and the target-dominant domains, as shown in Fig.~\ref{fig:overall_framework}. These two intermediate domains act as augmented views for the source and target domains in contrastive learning. Moreover, they are designed in a way that preserves the semantics of the dominant domain while learning the temporal characteristics of the less-dominant domain. Subsequently, we leverage in-domain contrastive learning to maximize the similarity between the source and the source-dominant domains, as well as maximizing the similarity between the target and target-dominant domains. Unlike the previous works that directly push the target domain towards the source domain using distance metrics or adversarial training, our proposed approach can progressively map the source and target domains towards an intermediate domain.

To summarize, our main contributions are as follows:

\begin{itemize}
    \item We propose \cotmix, a novel contrastive learning-based framework for time-series UDA. \cotmix deploys contrastive loss solely based on our Temporal Mixup strategy, making a unique way of addressing the problem of domain shift for time series data.
    
    \item We propose a novel cross-domain temporal mixup, a simple, generic, and effective strategy to generate new augmented views for in-domain contrastive learning at both source and target domains sides. This operation aims to fit contrastive learning to serve the adaptation objective.
    
    \item We conduct extensive experiments on five real-world time-series domain adaptation datasets. The results show that our \cotmix significantly outperforms state-of-the-art UDA methods.
\end{itemize}


\section{Related Works}

\subsubsection{Unsupervised Domain Adaptation}

UDA has drawn wide attention as a solution to reduce the gap between source and target distributions in different visual applications. Some methods focused on matching the statistical distribution of embeddings to learn domain invariant representations. For example, DDC \cite{ddc} trained an adaptation layer to jointly optimize classification performance and domain invariance based on MMD. Also, HoMM \cite{HoMM} explored aligning higher-order statistics for domain matching. DSAN \cite{dsan} proposed local MMD to align relevant subdomain distributions. Last, the concept of manifold criterion was introduced as a distance measure to validate the distribution matching across domains \cite{8674784}.

On the other hand, most methods deployed adversarial training for UDA. For example, DANN \cite{DANN} proposed a gradient reversal layer. DIRT-T \cite{dirt} added iterative refinement training to improve the adversarial training. CDAN \cite{CDAN} applied the adversarial adaptation to the information conveyed from the classifier predictions. Wang et al. \cite{ijcai2020p440} proposed a re-weighted adversarial domain adaptation with a triplet loss on the confusing domain to leverage both source samples and pseudo-labeled target samples. Triplet loss has also been deployed in \cite{WANG2023108993} to adjust the weights of pair-wise samples in intra-domain and inter-domain. Finally, Xu et al. \cite{domain_mixup} included the mixup operation on instance- and feature-level to improve the adversarial training. 

Furthermore, some methods explored contrastive learning to enhance the performance of the discrepancy measure \cite{can,cda}. In addition, contrastive learning has been also deployed in domain generalization. For example, the original sample-to-sample relations were replaced with proxy-to-sample relations to enhance the impact of positive alignment \cite{Yao_2022_CVPR}. Also, contrastive learning was utilized along with self-training for gaze estimation \cite{gaze_est}. For video UDA, \cite{contrast_mix} contrast the embeddings of unlabeled videos at different speeds with a background mixing mechanism. While these UDA techniques are proposed for visual applications, the problem of unsupervised domain adaptation for time-series data remains relatively under-explored, which is our focus in this work.

\begin{figure*}
\centering
\includegraphics[width=\textwidth]{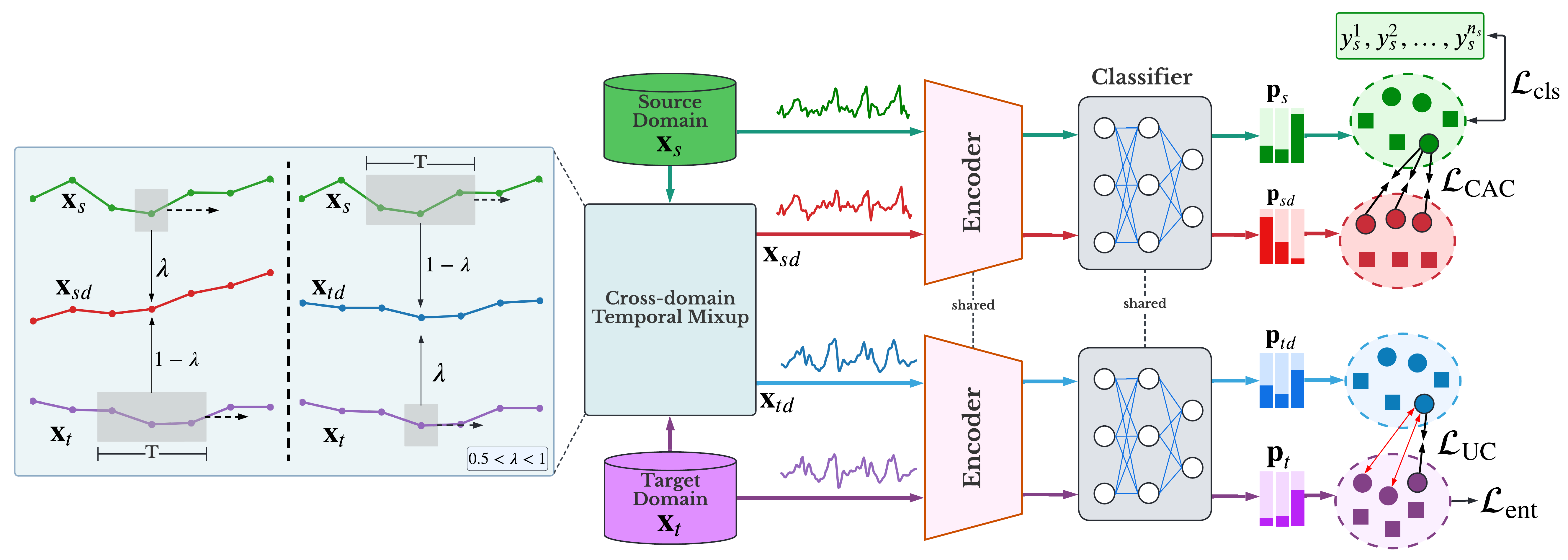}
    \caption{The overall structure of our CoTMix framework. The cross-domain temporal mixup strategy generates the source-dominant $\mathbf{x}_{sd}$, and the target-dominant $\mathbf{x}_{td}$ domains. We use a fixed $0.5 < \lambda < 1$ to ensure that one domain has more contribution in the generated samples. In addition, we aggregate $T$ timesteps from the less-dominant domain while the mixup to learn its temporal information. The generated domains act as augmented views for the in-domain contrastive learning, which is performed on the output probabilities.}
    \label{fig:overall_framework}
\end{figure*}

\subsubsection{Time-Series UDA Methods}

Few works were proposed for time-series UDA, despite its importance in many real-world applications. Some methods aimed to align the domains based on the feature statistical distribution. For example, AdvSKM \cite{dskn} reformed the MMD metric with a hybrid spectral kernel network to improve the general MMD metric. In addition, Net2Net ~\cite{quick_learning_tai} employed MMD loss between the new data and the previous one to serve as a quick learning mechanism for Fault diagnosis data. Also, TS-SASA \cite{aaai_ts_uda} aligned the intra- and inter-variable attention with MMD. The other methods deployed adversarial training used in visual UDA but with different architectures. For instance, VRADA \cite{vrada} used a variational recurrent neural network for feature extraction. Also, CoDATS \cite{codats} and CALDA \cite{calda} used DANN methodology with a 1D-CNN feature extractor and multi-layer fully connected layers. Some of these methods contributed to the design of the methodology.  For example, Jin et al. \cite{ts_da_attn} designed an unshared backbone, and a shared attention-based module to extract domain-specific information with adversarial training. In addition, SLARDA \cite{slarda} designed an autoregressive domain discriminator with a teacher model to align the class-wise distribution of the target domain. Also, Zhao et al.~\cite{EEG_UDA} implemented an adversarial UDA algorithm with center loss to constrain invariant feature space for EEG data. Finally, AdaTime \cite{adatime} was the first attempt to provide a benchmarking suite for time-series UDA, in which it unifies the backbone, datasets, and training schemes to ensure fair evaluation of time-series UDA methods.

Unlike these works that either follow discrepancy or adversarial training to move one domain towards the other, we propose to move each domain towards the other with a pure contrastive learning technique.

\subsubsection{Contrastive Learning}

The purpose of contrastive learning is to learn invariant representations based on data augmentation. Recently, many works have been proposed for self-supervised contrastive learning in visual applications \cite{simclr,moco,byol} and showed promising results. The success of these methods encouraged exploring contrastive learning for time-series data. For example, TNC \cite{tnc} exploited contrastive learning to ensure that neighboring timesteps are distinguishable from the non-neighboring timesteps. In addition, TS-TCC \cite{ts_tcc} proposed instance-wise temporal and contextual contrasting at the timestep level, while TS2VEC \cite{ts2vec} proposed contrastive learning in a hierarchical way for an arbitrary semantic level.

These methods rely on augmentations that can fit the representation learning tasks. However, they can not fit with UDA scenarios. Therefore, we propose a cross-domain temporal mixup strategy to generate augmented views that can be robust to the domain shift problem and narrow the gap between distant domains.


\section{Proposed Method}

\subsection{Problem Definition}
We address the problem of unsupervised domain adaptation for time-series data. Specifically, we have a labeled source domain $\mathcal{D}_s=\{(\mathbf{x}_{s}^i, y_{s}^i)\}_{i=1}^{n_s}$ with $n_s$ samples, and an unlabeled target domain $\mathcal{D}_t=\{\mathbf{x}_{t}^j\}_{j=1}^{n_t}$ with $n_t$ samples. Both domains have samples with a length of $L$ timesteps and share the same label space, i.e., $y_{s}^i, y_{t}^j \in \{1,2, \dots K\}$, where $K$ denotes the number of classes. It is assumed that there is a distribution shift between the two domains (i.e., $P(\mathbf{x}_s) \neq P(\mathbf{x}_t)$).
Given the source and target data, we aim to train a shared model that consists of a feature encoder  $\mathcal{F}(\cdot)$ and a classifier $\mathcal{C}(\cdot)$ to find a unified space that can successfully classify the unlabeled target data.

\subsection{Overview} 

In this section, we propose \cotmix, a \textbf{Co}ntrastive domain adaptation framework via \textbf{T}emporal \textbf{Mix}up for time-series data. \cotmix consists of two main components: the cross-domain temporal mixup strategy and the in-domain contrasting at the source and target sides. Fig.~\ref{fig:overall_framework} illustrates the overall structure of our proposed approach, which can be trained in an end-to-end manner.

\subsection{Temporal Mixup}

We propose a cross-domain temporal mixup strategy, in which we generate two new intermediate domains namely the source-dominant and the target-dominant domains using the mixup operation \cite{mixup}. Each of these domains should preserve the characteristics of one \textit{dominant} domain while considering the temporal information from the other \textit{less-dominant} domain. To do so, unlike the traditional mixup, we use a fixed mixup ratio $0.5 < \lambda < 1$, such that one domain will have more contribution than the other in the newly generated domain.
In addition, we learn the temporal information from the less-dominant domain by aggregating multiple forward and backward timesteps to be mixed with one timestep from the dominant domain, as illustrated in Fig.~\ref{fig:overall_framework}. For instance, we generate the source-dominant samples by mixing each timestep from the source domain with the average value of $T$ timesteps from the target domain ($\frac{T}{2}$ backward timesteps and $\frac{T}{2}$ forward timesteps), such that the source ratio is $\lambda$ and the target ratio is $1-\lambda$. We calculated the mean of timesteps as inspired by the moving average method~\cite{chatfield2013analysis}. Averaging timesteps has the advantages of eliminating short-term fluctuations and reducing the effect of extreme values.

Formally, given a source domain sample $x_s$ and a target domain sample $x_t$, we generate each timestep $i$ in the source-dominant domain as follows:
\begin{align}
    x_{sd}^{i} &= \lambda x_s^i + (1 - \lambda) \frac{1}{T} \sum_{j=i-\frac{T}{2}}^{i+\frac{T}{2}} x_t^j; \quad 0.5 < \lambda < 1 \label{eqn:src_mixup}
\end{align}
where $\mathbf{x}_{sd}=(x_{sd}^1, x_{sd}^2, \dots, x_{sd}^{L})$ represents the generated source-dominant sample, $T$ is the mixup window length, and $L$ is the sample length. 
A similar process is followed to generate the target-dominant samples, which can be formalized as follows:
\begin{align}
x_{td}^{i} &= \lambda x_t^i + (1 - \lambda) \frac{1}{T} \sum_{j=i-\frac{T}{2}}^{i+\frac{T}{2}} x_s^j, \quad 0.5 < \lambda < 1
    \label{eqn:trg_mixup} 
\end{align}
where $\mathbf{x}_{td}=(x_{td}^1, x_{td}^2, \dots, x_{td}^{L})$ represents the generated target-dominant sample.

\subsection{Contrastive Adaptation}

Given the generated source-dominant and target-dominant mixed domains, in addition to the original source and target domains, we use the feature encoder and the classifier to generate the probability vectors for the four domains. 
As inspired by \cite{li2021semantic}, we leverage the probability vectors in the InfoNCE loss \cite{cpc} to maximize the similarity between each domain and its corresponding intermediate view. Since we contrast each domain with its dominant mixed domain, we benefit from several advantages. First, we close up the gap between the two domains regardless of the shift distance, because the model keeps learning about the less-dominant domain on both sides \textit{progressively} throughout training. Second, in addition to mitigating the domain shift, in-domain contrastive learning improves the learning capability of the model about each domain separately.
Last but not least, this approach is simpler in implementation and training than traditional complex adversarial training approaches.

Fig.~\ref{fig:overall_framework} shows the overall structure of our framework. For the source domain side, we minimize the class-aware contrastive loss as well as the source classification loss. For the target domain side, we minimize the unsupervised contrastive loss and the entropy minimization loss. Next, we will discuss the losses on each side in more detail.

\subsubsection{Source Domain Side.}

Since we have access to the source domain labels, we leverage these labels to optimize both the in-domain class-aware contrastive loss and the standard cross-entropy loss.
The class-aware contrastive learning, as inspired by \cite{sup_con}, benefits from the available labeled data to include more positive pairs in the contrastive loss. In specific, for each anchor sample, we consider all the samples having the same class label within the mini-batch as positive pairs. In this way, we consider the semantic information between samples and avoid contrasting against false negatives, which could improve the quality of the learned representations. Moreover, by contrasting with multiple possible positive pairs (which are mixed with target domain data), we increase the chance of narrowing the gap between the anchor sample with samples having the same class label in the less-dominant domain (i.e., target domain), which further improves the class-wise alignment in the target domain.

Formally, given the $n_s$ source domain samples and the $n_s$ generated source-dominant samples, the overall samples become 2$n_s$. 
We generate the output probabilities $\mathbf{p}_s = \mathcal{C}(\mathcal{F}(\mathbf{x}_s))$ and $\mathbf{p}_{sd} = \mathcal{C}(\mathcal{F}(\mathbf{x}_{sd}))$. To this end, the overall source samples become ($\{\mathbf{x}_{so}^l, y_{so}^l\}_{l=1 \dots 2n_s}$), and their corresponding probabilities are $\mathbf{p}_{so}$. In addition, we assume that the class label is the same for any two corresponding samples from both domains, i.e., $y_{s}^i = y_{sd}^i$. Assuming that $k \in I \equiv \{1 \dots 2n_s\}$ represents the index of an arbitrary sample (from either source or source-dominant domains), and $A(k) \equiv I \setminus \{k\}$. The set of indices of all samples with the same class as an anchor sample $\mathbf{x}_{so}^{k}$ will be $U(k) = \{ u \in A(k) :  y_{so}^u = y_{so}^k\}$. Therefore, we can formulate the probabilistic class-aware contrasting loss $\mathcal{L}_{\mathrm{CAC}}$ as follows.

\begin{align}
    &\mathcal{L}_{\mathrm{CAC}}= \sum_{k \in I} \frac{-1}{|U(k)|} \sum_{u \in U(k)} \log \frac{\exp \left(\mathbf{p}_{so}^k \boldsymbol{\cdot} \mathbf{p}_{so}^u / \tau\right)}{\sum_{a \in A(k)} \exp \left(\mathbf{p}_{so}^k \boldsymbol{\cdot} \mathbf{p}_{so}^a / \tau\right)}, \label{eqn:src_sup_con}
\end{align}

where $\boldsymbol{\cdot}$ symbol denotes the inner dot product, $\tau$ is a temperature parameter, and $|U(k)|$ is the cardinality of $U(k)$.

In addition to the class-aware contrasting, we also train the model to minimize the cross-entropy loss as follows.

\begin{align}
    &\mathcal{L}_{\mathrm{cls}}= -\mathbb{E}_{(\mathbf{x}_{s},y_{s}) \sim P_{s}} \sum_{k=1}^K  \mathbbm{1}_{[y_s = k]} \log \mathbf{p}_{s}(k).
    \label{eqn:cross-entropy}
\end{align}

\subsubsection{Target Domain Side.}
Considering the target domain, we do not have access to its label information. Therefore, we can only contrast its samples in an unsupervised manner. Given the $n_t$ target domain samples and its corresponding $n_t$ target-dominant mixed domain samples, then the total number of samples becomes $2n_t$. We calculate the output probabilities $\mathbf{p}_t = \mathcal{C}(\mathcal{F}(\mathbf{x}_t))$ and $\mathbf{p}_{td} = \mathcal{C}(\mathcal{F}(\mathbf{x}_{td}))$. Therefore, the overall target samples become ($\{\mathbf{x}_{to}^l\}_{l=1 \dots 2n_t}$), and their corresponding probabilities are $\mathbf{p}_{to}$, such that for each target sample $\mathbf{x}_t^i$, $1\leq i \leq n_t$, it forms a positive pair with its corresponding target-dominant sample $\mathbf{x}_{td}^i$ and vice versa.
Assuming that $\mathbf{p}_{to} = (\mathbf{p}_{t}^1, \mathbf{p}_{t}^2, \dots,  \mathbf{p}_{t}^{n_t}, \mathbf{p}_{td}^{1}, \dots, \mathbf{p}_{td}^{n_t})$, where $\mathbf{p}_{to}^{k} = \mathbf{p}_{t}^{k}$ if $k \leq n_t$, and $\mathbf{p}_{to}^{k} = \mathbf{p}_{td}^{k-n_t}$ otherwise. For an anchor sample indexed $k \in I \equiv \{1 \dots 2n_t\}$, $A(k) \equiv I \setminus \{k\}$, we define $f(k)$ as the index of the positive pair of $k$,  such that $f(k) = k + n_t$ if $k \leq n_t$, and $f(k) = k - n_t$ otherwise.
Next, we design the probabilistic unsupervised contrastive loss $\mathcal{L}_{\mathrm{UC}}$ as follows.

\begin{align}
    \mathcal{L}_{\mathrm{UC}} &= \frac{-1}{2n_t}\sum_{k \in I} \log \frac{\exp \left(\mathbf{p}_{to}^k \boldsymbol{\cdot} \mathbf{p}_{to}^{f(k)} / \tau\right)}{\sum_{a \in A(k)} \exp \left(\mathbf{p}_{to}^k \boldsymbol{\cdot} \mathbf{p}_{to}^a / \tau\right)}. \label{eqn:trg_con}
\end{align}

In addition, we minimize the entropy on the unlabeled target domain, formulated as follows.

\begin{align}
    &\mathcal{L}_{\mathrm{ent}}= -\mathbb{E}_{\mathbf{x}_{t}\sim P_{t}} [ \mathbf{p}_{t}^\top \log \mathbf{p}_{t}].
    \label{eqn:entropy}
\end{align}

Leveraging this entropy minimization loss forces the classifier to be confident about its prediction for the target domain data \cite{dirt}.

\subsection{Overall Objective}
Our proposed \cotmix is trained with a simple procedure. 
First, we mix the input signals from the source and target domains to generate the source-dominant and the target-dominant mixed samples as in Equations~\ref{eqn:src_mixup} and \ref{eqn:trg_mixup}, respectively. 
Next, we develop the class-aware contrastive training in the source domain side, and the unsupervised contrastive training in the target domain side as in Equations~\ref{eqn:src_sup_con} and \ref{eqn:trg_con}, to minimize $\mathcal{L}_{\text{CAC}}$ and $\mathcal{L}_{\text{UC}}$ respectively. 
Besides minimizing the contrastive losses, we also minimize the standard cross-entropy loss $\mathcal{L}_{\text{cls}}$ on the labeled source domain, as well as the conditional entropy loss $\mathcal{L}_{\text{ent}}$ on the unlabeled target domain.
Overall, the training objective is to minimize these losses combined as follows.
\begin{align}
    \min_{\mathcal{F}, \mathcal{C}} \mathcal{L}_{\text{overall}} = \underbrace{\beta_1 \mathcal{L}_{\mathrm{cls}} + \beta_2 \mathcal{L}_{\mathrm{CAC}}}_{\text{Source side}} + \underbrace{\beta_3 \mathcal{L}_{\mathrm{ent}} +\beta_4  \mathcal{L}_{\mathrm{UC}}}_{\text{Target side}},
\label{eqn:overall}
\end{align}
where the hyperparameters $\beta_1$, $\beta_2$, $\beta_3$, and $\beta_4$ control the contribution of each loss on the overall performance.


\section{Experimental Setup}

\subsection{Datasets}
To evaluate our proposed approach, we choose five real-world time-series datasets in three applications, i.e., sleep stage classification, human activity recognition, and fault detection. These datasets have different characteristics in terms of complexity, the number of samples, the sample length and type, the number of sensors, and the severity of the domain shift. In the first four datasets, we treat data from each subject as a separate domain, since different subjects may have distinct behaviors, leading to distribution shifts. 

The first dataset is \textbf{SSC} for sleep stage classification, and it includes classifying electroencephalography (EEG) signals into one of five classes, i.e., Wake (W), Non-Rapid Eye Movement (N1, N2, N3), and Rapid Eye Movement (REM). We selected a single EEG channel (i.e., Fpz-Cz) from the sleep-EDF \cite{sleepEDF_dataset} dataset, following previous studies \cite{attnSleep_paper}.

The second dataset is \textbf{UCIHAR} for Human Activity Recognition \cite{uciHAR_dataset}. It includes three sensors' readings i.e., accelerometer, gyroscope, and body sensors, collected from 30 subjects. 

The third dataset is \textbf{HHAR} (Heterogeneity Human Activity Recognition) dataset \cite{hhar_dataset}, and it was collected from 9 different users using smartphones and smartwatches. We consider the data from the Samsung smartphone following \cite{adatime}. 

\input{tables/datasets}

Forth, the \textbf{WISDM} dataset \cite{wisdm_dataset} is also for human activity recognition, and it was collected with accelerometer sensors from 36 subjects. 
The objective in the latter three datasets is to classify sensors' readings into one of six activities, i.e., walking, walking upstairs, downstairs, standing, sitting, and lying down. 

Last, \textbf{Boiler} fault detection dataset \cite{aaai_ts_uda} describes three boilers, where each one is considered as a separate domain. The objective is to help predicting the faulty blowdown valve of each boiler. The data is imbalanced because obtaining faulty samples in the mechanical system is hard.

Since each dataset contains numerous subjects (i.e., domains), we selected five random scenarios as in \cite{codats,dskn}. Each scenario represents the ID of the source subject (domain) and the ID of the target subject (domain) within their respective datasets. For example, scenario 16 $\rightarrow$ 1 in the SSC dataset indicates that the source is subject \#16 and the target is subject \#1.
More details about the datasets are illustrated in Table~\ref{tbl:datasets}. We included three large-scale datasets, i.e., SSC, HHAR, and Boiler, and two small-scale datasets, i.e., UCIHAR and WISDM datasets. This testifies to the capability of our proposed framework to adapt different scales of datasets/domains.

\input{tables/results_dev_risk}

\subsection{Baselines}
We compared our proposed method with seven state-of-the-art UDA methods that span both discrepancy- and adversarial-based schemes as follows: 
\begin{itemize}
    \item \textbf{FCN}: Fully Convolutional Network, representing the source-only experiment.
    \item \textbf{HoMM}~\cite{HoMM}: Higher-order Moment Matching.
    \item \textbf{DSAN}~\cite{dsan}: Deep Subdomain Adaptation.
    \item \textbf{DANN}~\cite{DANN}: Domain-Adversarial Training of Neural Networks.
    \item \textbf{CDAN}~\cite{CDAN}: Conditional Domain Adversarial Network for Adaptation.
    \item \textbf{DIRT-T}~\cite{dirt}: Decision-boundary Iterative Refinement Training with a Teacher.
    \item \textbf{CoDATS}~\cite{codats}: Convolutional deep Domain Adaptation model for Time Series.
    \item \textbf{AdvSKM}~\cite{dskn}: Adversarial Spectral Kernel Matching.
\end{itemize}


\subsection{Implementation Details}
\subsubsection{Dataset preprocessing}
We split each domain into 70/30\%, where the 70\% splits in both domains are for training, while the 30\% in the source domain is treated as a validation set for the risk calculation, and the 30\% in the target domain acts as a test set. In addition, all the splits were normalized based on the training statistics \cite{codats}. We applied a sliding window of 128 for the three human activity recognition datasets, but for the SSC dataset, we kept the original sample length of 3000 timesteps.

%
\begin{figure}[!htb]
    \centering
    \includegraphics[width=1.05\columnwidth]{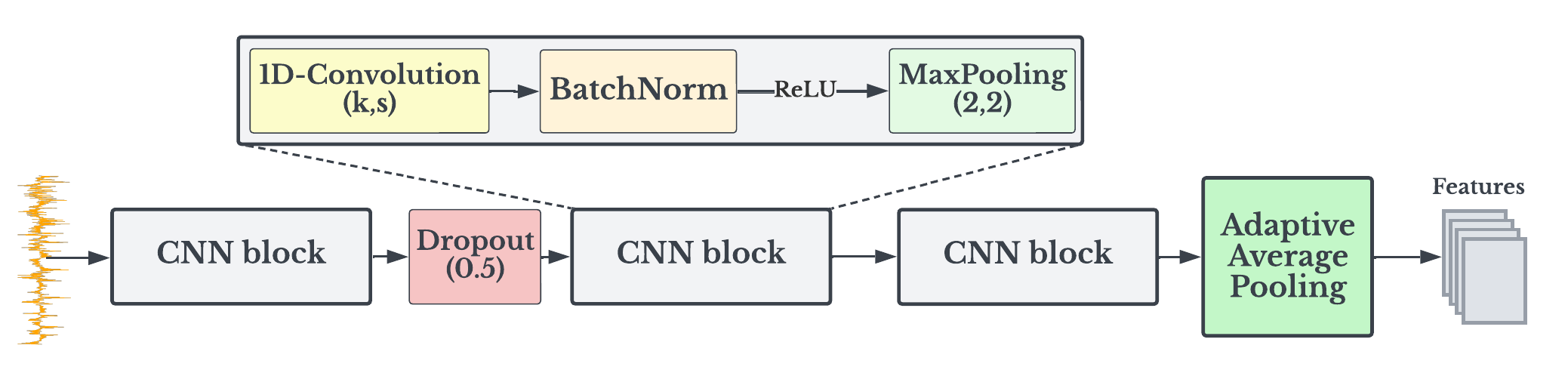}
    \caption{Feature encoder in our CoTMix framework. The pair ($k,s$) refers to the kernel size and stride respectively.}
    \label{fig:fe}
\end{figure}

\subsubsection{Feature encoder}
We adopted a convolutional neural network as our feature encoder~\cite{ts_tcc,catcc}. As shown in Fig.~\ref{fig:fe}, our feature encoder consists of three similar convolutional blocks. The first one is followed by a dropout layer, while the last one is followed by an adaptive average pooling layer. Each block contains a 1D-convolutional layer, a batch normalization layer, a non-linearity ReLU activation function, and a 1D-maxpooling layer. The filter sizes of the 1D-convolution layers are set to 64, 128, and 128 in the three blocks respectively. The kernel size and stride values differ from one dataset to the other. We set this pair as (5,1) for UCIHAR, HHAR, and WISDM datasets as they have the same sequence length $L$. For the SSC dataset, we set it as (25,6) due to its longer sequence length. This encoder is followed by a single fully connected layer for classification.

\subsubsection{Unified training scheme}
To ensure a fair evaluation, we unified the way of training, the backbone encoder, the hyperparameters search methodology, and the risk minimization setting for our proposed approach as well as all the baselines, as inspired by AdaTime benchmark \cite{adatime}. In specific, we trained all the models for 40 epochs with a batch size of 32 and optimized the neural network weights using Adam optimizer with a learning rate of 1e-3. The reported results show the average and the standard deviation performance of the last epoch of training, for three repeated experiments with three different seeds. 

\input{tables/selected_hyper}

To choose the hyperparameters of our \cotmix, i.e., $\lambda$, $\beta_1, \beta_2, \beta_3, \beta_4$, as well as the hyperparameters of the baselines, we performed  a hyperparameter sweep with 100 trials. The selection of these hyperparameters was from predefined ranges using uniform sampling. Based on this hyperparameters search, we picked the best model that minimizes the realistic Deep Embedded Validation (DEV) risk \cite{dev_risk, adatime}. This risk does not consider any target labels to be calculated. Instead, it considers the highly correlated source features to the target features via importance weighting schemes, which give lower weights to the less correlated features. Despite that choosing the hyperparameters based on this risk may not yield the best performance on the target domain, however, it ensures a fair and realistic evaluation scheme and prevents overestimated results.
We included the ranges, as well as the values of the selected hyperparameters, in Table~\ref{tbl:selected_hpr_values}.

\input{Figures/diff_mixups}


\section{Results}

\subsection{Comparison with Baselines}
Table~\ref{tbl:all_results_dev_risk} reports the macro-F1 (MF1) scores of our proposed framework against the other competing state-of-the-art methods on the five benchmark datasets. The MF1-score metric is more suitable to reflect the true performance of the imbalanced time-series data. The results show that our proposed \cotmix approach outperforms adversarial and discrepancy baselines significantly in the overall performance across the five datasets, indicating its effectiveness. For the SSC dataset, it achieves a 2.65\% improvement over the second-best baseline. For UCIHAR, HHAR, and WISDM datasets, it was able to achieve 3.51\%, 5.59\%, and 5.09\% improvement over the second-best method. Last, for the challenging Boiler dataset, it improves by 0.25\% over the HoMM method. Since the temporal mixup operation is being performed on the input space, we notice more performance improvement in the human activity recognition datasets, which have less complex time-series data, compared to the more complex SSC and Boiler datasets.

Additionally, our \cotmix shows a significant improvement in big domain shifts. For example, in the HHAR dataset, we find that scenario 2$\rightarrow$7 suffers a big shift, indicated by the poor source-only performance of 38.03\%. We find that \cotmix improved its performance by 30.73\% reaching 68.76\%. Despite that \cotmix achieves the best average performance over the baselines, it can still achieve less performance than some baselines on some cross-domain scenarios. For example, in the HHAR dataset, DSAN shows a better performance than \cotmix in three cross-domain scenarios. However, the difference is not significant with a maximum of 2.45\% in the worst case, and both methods are already close to 100\%. Counterpart, \cotmix outperforms DSAN by 23.15\% in one cross-domain scenario, which explains the improved average performance. Similarly, for other datasets, if \cotmix is not the best-performing method in one cross-domain scenario, the gap is not significant, except for only a single scenario in the WISDM dataset.

To test the effectiveness of the class-aware contrastive loss on the source domain side, we added a second variant of our framework (CoTMix*), in which we deploy the unsupervised contrastive loss (Equation~\ref{eqn:trg_con}) on the source domain side. We notice that it consistently achieves less performance than \cotmix, however, it always grades the second-best average performance outperforming other baselines. This indicates the efficacy of our method and also shows that considering the semantic information while contrasting helps to improve the class-conditional alignment in the target domain.


\subsection{Study of Different Augmentations}

In this section, we compare the capability of our proposed temporal mixup to mitigate the domain shift against other augmentations proposed for time-series representation learning tasks. In specific, we replaced our cross-domain temporal mixup with four different augmentations, i.e., permutation, scaling, jittering \cite{ts_tcc}, and masking \cite{ts2vec}. The experimental results are provided in Table~\ref{tbl:diff_aug_exp}.
It can be clearly noticed that our proposed temporal mixup strategy is more effective than other augmentations in the UDA settings. These augmentations may enhance the in-domain representation learning capability of the model, but they do not serve the adaptation objective, i.e., reducing the domain shift. This shows how our temporal mixup contributes to the success of contrastive learning for domain adaptation.


\subsection{Mixup Strategies}
In our temporal mixup, we use a fixed mixup ratio $0.5 < \lambda < 1$ to keep the semantic characteristics of one domain for in-domain contrastive learning. Nevertheless, we compare using this fixed mixup ratio with two different strategies. The first is the random mixup ratio selected randomly from a beta distribution $\lambda \sim Beta(\alpha, \alpha)$ as in the traditional mixup \cite{mixup}. The second is to specify $\lambda$ from a range by randomly selecting it from the beta distribution and limiting it to a specific range \cite{fixbi}. In specific, $\lambda' \sim max(\lambda, 1-\lambda)$, where $\lambda \sim Beta(\alpha, \alpha)$.

\input{tables/other_augmentations}

Fig.~\ref{fig:diff_mixups} shows the comparison results for the three scenarios, where we report the average performance of three experiments at each point in the sub-figures. In general, we find that the ``Random" mixup strategy causes noticeable performance degradation, as it does not ensure keeping most of the semantics of one dominant domain while contrastive learning. On the other hand, the ``Range" mixup strategy keeps the mixup ratio within a range $\geq 0.5$, which ensures having a more dominant domain, but with different random ratios. Therefore, it achieves a better performance than the ``Random" mixup strategy, but its randomness affects the performance. Finally, we find that using a fixed mixup ratio can achieve the best performance, as it ensures stable ratios of the domains in the augmented views.

Notably, this analysis shows that we may achieve better results than those viewed in Table~\ref{tbl:all_results_dev_risk}. The reason is that these values are selected by minimizing the DEV risk \cite{dev_risk, adatime}. This risk may not provide the optimal performance on the target data for some datasets, however, it is more realistic in real-world scenarios. In Table~\ref{tbl:risks_comparison}, we show that our \cotmix can achieve better performance if hyperparameters were selected based on the target risk, i.e., based on the labels of the target domain.

\input{tables/risks_comparison}


\input{tables/ablation}

\subsection{Ablation Study}
Since our \cotmix combines three losses in addition to $\mathcal{L}_{\text{cls}}$, we study the effect of these losses on the overall performance to provide additional insights on what makes \cotmix performant. Table~\ref{tbl:ablation} presents this ablation study, where $\mathcal{L}_{\text{cls}}$ is present by default in all the cases. In particular, we first omit all three losses, to show the bottom-line performance. Next, we add the entropy minimization loss to the training.
After that, we apply contrastive training on only one side interchangeably. Finally, we show the results with all the losses together.

We arrive at two conclusions. First, adding entropy minimization improves the overall performance, as it helps the classifier to be more confident about the unlabeled target domain. Second, applying the contrastive loss to only one side still improves the performance. This indicates that moving one domain towards an intermediate domain by considering the cross-domain temporal relations is still effective for adaptation. Moreover, we find that SSC and UCIHAR datasets achieve better performance by contrasting only on the source side than contrasting only on the target side. Counterpart, the performance on HHAR and WISDM datasets improve more with contrasting only on the target side than the source side. This can be regarded to the efficacy of the learned temporal features from one side over the other for adapting the two domains. This experiment also shows that moving only one domain towards the intermediate domain may not be the most effective way. Nevertheless, the performance is consistently the best when contrastive losses are applied on both sides, i.e., moving both domains.

\input{Figures/sens_figure}

\subsection{Selection of Temporal Mixup Window}
\label{sec:temporal_mixup_value}
Our proposed temporal mixup could be affected by the length of aggregated timesteps $T$ in the less-dominant domain. Therefore, we study its significance on the performance, as shown in Fig.~\ref{Fig:sens_overall}. To avoid random selection to the value of $T$, we assign it as a proportion in the sample length $L$. The first case in the analysis, (i.e., $T=0$) represents a one-to-one mixup of timesteps among source and target domains, while in the next cases, we gradually increase $T$.

First, we notice that when $T=0$, which corresponds to the traditional mixup~\cite{mixup}, the performance usually drops. This shows the significance of our temporal mixup strategy, as it significantly improves the performance in the four datasets. We find that the best performance is achieved with $T=0.1L$ in UCIHAR and HHAR datasets, and $T=0.05L$ in SSC and WISDM datasets. Increasing $T$ beyond these values can still achieve better performance than the usual mixup until some point, where the performance is then hurt. This could be regarded to transferring more irrelevant information than the proper window of temporal information. Based on this sensitivity analysis, we recommend searching for the best value $T$ in the interval $[0.05L, 0.2L]$.


\section{Conclusions}
In this work, we propose a novel time-series unsupervised domain adaptation framework (\cotmix) that exploits contrastive learning to mitigate the domain shift problem. Specifically, we develop a cross-domain temporal mixup to generate augmented views in both source and target domain sides and then leverage these augmented views for in-domain contrasting. The extensive experiments proved the superiority of our proposed approach over state-of-the-art UDA methods. In addition, the ablation study showed the importance of contrastive learning in both source and target domain sides to narrow down the domain shift. Finally, we show that unlike other augmentations proposed for representation learning tasks, our cross-domain temporal mixup is more robust against the domain shift. 

\balance
\bibliographystyle{unsrt}
\bibliography{references}

\end{document}

%% file: tables/datasets.tex
\begin{table}[!t]
\centering
\caption{Details of the adopted datasets (C: \#channels, K: \#classes, L: sample length).}
\begin{NiceTabular}{l|ccc|cc}
\toprule
\textbf{Dataset} & \textbf{C}  & \textbf{K} & \textbf{L} & \# training samples & \# testing samples \\ \midrule
SSC & 1 & 5 & 3000 & 14280 & 6130 \\ 
UCIHAR &  9 & 6 & 128 & 2300 & 990 \\ 
HHAR & 3 & 6 & 128 & 12716 & 5218 \\ 
WISDM  & 3 & 6 & 128 & 1350 & 720 \\
Boiler & 20 & 2 & 36 & 26785 & 44687 \\
\bottomrule
\end{NiceTabular}
\label{tbl:datasets}
\end{table}

%% file: tables/results_dev_risk.tex
\begin{table*}[!tb]
\centering
\caption{Detailed results of each cross-domain scenario in the five adopted datasets in terms of MF1 score. CoTMix* indicates deploying unsupervised contrastive loss in the source side. The results are based on minimizing the DEV risk. The \textbf{best} results are in bold, and the \underline{second-best} results are underlined.}
\resizebox{\textwidth}{!}{
\begin{NiceTabular}{@{}c|c|cccccccc|cc@{}}
\toprule
\textbf{Dataset} & \textbf{Scenario} & \textbf{FCN} & \textbf{HoMM} & \textbf{DSAN} & \textbf{DANN} & \textbf{CDAN} & \textbf{DIRT-T} & \textbf{CoDATS} & \textbf{AdvSKM} & \textbf{CoTMix*} &\textbf{\cotmix} \\ \midrule


& 16$\rightarrow$1 & 52.44\scriptsize$\pm$4.78 & 55.57\scriptsize$\pm$2.00 & 58.76\scriptsize$\pm$2.02 & 58.78\scriptsize$\pm$4.76 & \underline{60.95\scriptsize$\pm$1.13} & 54.4\scriptsize$\pm$12.46 & 60.03\scriptsize$\pm$1.18 & 57.80\scriptsize$\pm$0.69 & 59.85\scriptsize$\pm$4.39 & \textbf{60.97\scriptsize$\pm$4.32} \\
& 9$\rightarrow$14 & 64.35\scriptsize$\pm$5.66 & 63.66\scriptsize$\pm$1.48 & \underline{69.45\scriptsize$\pm$4.04} & 64.61\scriptsize$\pm$0.93 & 60.5\scriptsize$\pm$10.01 & \textbf{71.33\scriptsize$\pm$3.72} & 52.2\scriptsize$\pm$10.55 & 64.27\scriptsize$\pm$2.93 & 65.48\scriptsize$\pm$1.45 & 65.80\scriptsize$\pm$3.15 \\
& 12$\rightarrow$5 & 56.07\scriptsize$\pm$3.07 & 55.87\scriptsize$\pm$2.93 & 64.92\scriptsize$\pm$1.65 & \textbf{65.47\scriptsize$\pm$0.95} & 65.01\scriptsize$\pm$1.34 & \underline{64.99\scriptsize$\pm$4.98} & 56.96\scriptsize$\pm$2.41 & 55.12\scriptsize$\pm$2.52 & 64.03\scriptsize$\pm$4.47 & 61.59\scriptsize$\pm$1.98 \\
& 7$\rightarrow$18 & 59.39\scriptsize$\pm$4.94 & 67.49\scriptsize$\pm$1.51 & 68.69\scriptsize$\pm$0.99 & 68.88\scriptsize$\pm$2.81 & 67.02\scriptsize$\pm$1.13 & \underline{69.94\scriptsize$\pm$0.43} & 68.64\scriptsize$\pm$2.93 & 67.31\scriptsize$\pm$3.83 & 64.79\scriptsize$\pm$1.30 & \textbf{73.34\scriptsize$\pm$1.27} \\
& 0$\rightarrow$11 & 43.80\scriptsize$\pm$5.83 & 50.93\scriptsize$\pm$4.31 & 37.43\scriptsize$\pm$2.92 & 31.13\scriptsize$\pm$1.74 & 30.8\scriptsize$\pm$10.69 & 35.62\scriptsize$\pm$3.79 & 41.12\scriptsize$\pm$5.14 & \textbf{55.11\scriptsize$\pm$4.56} & 48.82\scriptsize$\pm$4.60 & \underline{51.16\scriptsize$\pm$3.71} \\
\cmidrule{2-12}
\multirow{-6}{*}{SSC} & AVG & 55.21 & 58.70 & 59.85 & 57.77 & 56.86 & 59.26 & 55.79 & 59.92 & \underline{60.59} & \textbf{62.57} \\

\bottomrule
\toprule
& 2$\rightarrow$11 & 58.39\scriptsize$\pm$3.87 & 73.38\scriptsize$\pm$7.34 & 75.58\scriptsize$\pm$9.18 & 77.8\scriptsize$\pm$18.26 & 71.51\scriptsize$\pm$8.84 & 88.44\scriptsize$\pm$9.23 & 51.81\scriptsize$\pm$4.67 & 65.74\scriptsize$\pm$2.69 & \textbf{99.01\scriptsize$\pm$0.71} & \underline{97.17\scriptsize$\pm$4.00} \\
& 12$\rightarrow$16 & 56.06\scriptsize$\pm$2.80 & 59.84\scriptsize$\pm$1.43 & 61.71\scriptsize$\pm$1.75 & 63.26\scriptsize$\pm$2.49 & 54.66\scriptsize$\pm$2.91 & 58.47\scriptsize$\pm$2.98 & 54.81\scriptsize$\pm$2.76 & 60.09\scriptsize$\pm$1.40 & \underline{66.59\scriptsize$\pm$5.53} & \textbf{77.56\scriptsize$\pm$1.34} \\
& 9$\rightarrow$18 & 58.87\scriptsize$\pm$5.93 & 60.0\scriptsize$\pm$11.83 & 67.10\scriptsize$\pm$4.61 & 57.49\scriptsize$\pm$7.77 & 40.94\scriptsize$\pm$3.18 & 65.9\scriptsize$\pm$13.25 & 31.83\scriptsize$\pm$8.89 & 53.70\scriptsize$\pm$4.61 & \underline{68.46\scriptsize$\pm$5.64} & \textbf{75.29\scriptsize$\pm$5.62} \\
& 6$\rightarrow$23 & 43.59\scriptsize$\pm$8.34 & 90.48\scriptsize$\pm$0.80 & \underline{93.22\scriptsize$\pm$2.49} & \textbf{95.86\scriptsize$\pm$1.84} & 61.31\scriptsize$\pm$9.02 & 90.56\scriptsize$\pm$8.73 & 81.23\scriptsize$\pm$4.07 & 79.31\scriptsize$\pm$8.95 & 90.94\scriptsize$\pm$2.78 & 91.74\scriptsize$\pm$3.29 \\
& 7$\rightarrow$13 & 87.45\scriptsize$\pm$6.20 & 85.94\scriptsize$\pm$2.52 & 88.82\scriptsize$\pm$3.08 & \underline{91.71\scriptsize$\pm$0.84} & 82.1\scriptsize$\pm$11.91 & \textbf{93.73\scriptsize$\pm$0.56} & 80.9\scriptsize$\pm$13.74 & 88.89\scriptsize$\pm$3.12 & 89.80\scriptsize$\pm$1.66 & 88.47\scriptsize$\pm$3.27 \\
\cmidrule{2-12}
\multirow{-6}{*}{UCIHAR} & AVG & 60.87 & 78.28 & 81.07 & 80.89 & 64.66 & 82.54 & 65.12 & 74.62 & \underline{82.96} & \textbf{86.05} \\

\bottomrule
\toprule
& 0$\rightarrow$6 & 54.26\scriptsize$\pm$3.46 & 63.58\scriptsize$\pm$2.24 & 58.81\scriptsize$\pm$7.19 & 46.54\scriptsize$\pm$0.61 & 45.52\scriptsize$\pm$0.94 & 52.63\scriptsize$\pm$9.77 & 44.73\scriptsize$\pm$1.65 & 45.52\scriptsize$\pm$0.91 & \underline{66.81\scriptsize$\pm$1.79} & \textbf{69.34\scriptsize$\pm$4.66} \\
& 1$\rightarrow$6 & 64.09\scriptsize$\pm$4.02 & 88.49\scriptsize$\pm$2.00 & \textbf{93.42\scriptsize$\pm$0.64} & 90.73\scriptsize$\pm$1.97 & 92.99\scriptsize$\pm$0.70 & \underline{93.10\scriptsize$\pm$2.06} & 91.98\scriptsize$\pm$1.01 & 92.99\scriptsize$\pm$0.72 & 90.49\scriptsize$\pm$0.68 & 91.97\scriptsize$\pm$2.01 \\
& 2$\rightarrow$7 & 38.03\scriptsize$\pm$4.42 & 47.12\scriptsize$\pm$4.27 & 45.61\scriptsize$\pm$0.51 & 46.58\scriptsize$\pm$3.13 & 54.12\scriptsize$\pm$7.12 & \underline{63.49\scriptsize$\pm$1.95} & 47.56\scriptsize$\pm$5.04 & 54.11\scriptsize$\pm$7.12 & 62.16\scriptsize$\pm$0.41 & \textbf{68.76\scriptsize$\pm$8.54} \\
& 3$\rightarrow$8 & 79.40\scriptsize$\pm$1.44 & 79.23\scriptsize$\pm$1.13 & \textbf{98.44\scriptsize$\pm$0.23} & 83.4\scriptsize$\pm$10.12 & \underline{98.17\scriptsize$\pm$0.37} & 87.1\scriptsize$\pm$10.06 & 91.83\scriptsize$\pm$4.56 & \underline{98.17\scriptsize$\pm$0.37} & 95.74\scriptsize$\pm$0.81 & 96.01\scriptsize$\pm$0.85 \\
& 4$\rightarrow$5 & 78.75\scriptsize$\pm$4.09 & 84.07\scriptsize$\pm$1.19 & \textbf{98.47\scriptsize$\pm$0.32} & 95.83\scriptsize$\pm$0.28 & 96.39\scriptsize$\pm$1.37 & \underline{97.13\scriptsize$\pm$0.44} & 92.52\scriptsize$\pm$3.14 & 96.39\scriptsize$\pm$1.37 & 97.03\scriptsize$\pm$0.62 & 96.61\scriptsize$\pm$1.70 \\
\cmidrule{2-12}
\multirow{-6}{*}{HHAR} & AVG & 62.90 & 72.50 & 78.95 & 72.62 & 77.43 & 78.69 & 73.72 & 77.43 & \underline{82.45} & \textbf{84.54} \\


\bottomrule
\toprule
& 35$\rightarrow$31 & 40.98\scriptsize$\pm$7.60 & \textbf{66.29\scriptsize$\pm$0.84} & 57.25\scriptsize$\pm$6.07 & 52.21\scriptsize$\pm$1.09 & 49.02\scriptsize$\pm$4.20 & 46.75\scriptsize$\pm$3.54 & 40.96\scriptsize$\pm$19.0 & \underline{61.91\scriptsize$\pm$6.95} & 48.77\scriptsize$\pm$3.15 & 47.68\scriptsize$\pm$6.97 \\
& 7$\rightarrow$18 & 35.25\scriptsize$\pm$4.87 & 48.67\scriptsize$\pm$6.31 & 52.77\scriptsize$\pm$2.23 & 41.16\scriptsize$\pm$6.62 & 57.65\scriptsize$\pm$0.18 & \underline{57.89\scriptsize$\pm$0.15} & 42.00\scriptsize$\pm$3.75 & 49.84\scriptsize$\pm$5.31 & 56.59\scriptsize$\pm$2.87 & \textbf{74.88\scriptsize$\pm$2.75} \\
& 20$\rightarrow$30 & 61.52\scriptsize$\pm$2.14 & 65.28\scriptsize$\pm$2.45 & 63.39\scriptsize$\pm$0.70 & 71.98\scriptsize$\pm$10.1 & 65.50\scriptsize$\pm$0.61 & 65.49\scriptsize$\pm$0.62 & 69.65\scriptsize$\pm$7.60 & 69.35\scriptsize$\pm$1.38 & \textbf{90.64\scriptsize$\pm$0.14} & \underline{77.90\scriptsize$\pm$1.78} \\
& 6$\rightarrow$19 & 49.09\scriptsize$\pm$4.34 & 63.78\scriptsize$\pm$4.35 & 53.35\scriptsize$\pm$5.37 & 59.09\scriptsize$\pm$3.57 & 44.03\scriptsize$\pm$0.81 & 45.16\scriptsize$\pm$0.00 & \textbf{70.6\scriptsize$\pm$12.51} & 54.89\scriptsize$\pm$4.14 & 56.74\scriptsize$\pm$8.75 & \underline{64.27\scriptsize$\pm$1.99} \\
& 18$\rightarrow$23 & 49.55\scriptsize$\pm$9.12 & 62.11\scriptsize$\pm$7.57 & 55.76\scriptsize$\pm$1.46 & 48.00\scriptsize$\pm$0.90 & 50.16\scriptsize$\pm$0.44 & 50.89\scriptsize$\pm$0.40 & 48.2\scriptsize$\pm$15.11 & 51.3\scriptsize$\pm$10.33 & \textbf{67.99\scriptsize$\pm$4.24} & \underline{66.87\scriptsize$\pm$2.02} \\
\cmidrule{2-12}
\multirow{-6}{*}{WISDM} & AVG & 47.28 & 61.23 & 56.51 & 54.48 & 53.27 & 53.24 & 54.27 & 57.46 & \underline{64.15} & \textbf{66.32} \\ 

\bottomrule
\toprule
& 1$\rightarrow$2 
& 49.09\scriptsize$\pm$0.45 
& \textbf{51.09\scriptsize$\pm$1.70} 
& 47.13\scriptsize$\pm$0.36 
& 46.81\scriptsize$\pm$0.36 
& 47.97\scriptsize$\pm$0.86 
& 47.97\scriptsize$\pm$1.20 
& 47.22\scriptsize$\pm$0.10 
& 49.69\scriptsize$\pm$0.68 
& 45.90\scriptsize$\pm$0.03 
& \underline{50.84\scriptsize$\pm$1.62} \\
& 1$\rightarrow$3
& 70.07\scriptsize$\pm$9.02
& 90.41\scriptsize$\pm$1.46
& 91.61\scriptsize$\pm$1.86
& 90.53\scriptsize$\pm$0.95
& 88.12\scriptsize$\pm$1.51
& 61.46\scriptsize$\pm$11.6
& 89.58\scriptsize$\pm$1.75 
& \underline{91.72\scriptsize$\pm$1.09} 
& 91.41\scriptsize$\pm$0.88 
& \textbf{92.37\scriptsize$\pm$1.01} \\
& 3$\rightarrow$1
& 63.04\scriptsize$\pm$3.25
& 49.05\scriptsize$\pm$0.08
& 47.92\scriptsize$\pm$0.00
& 48.97\scriptsize$\pm$0.34
& 48.91\scriptsize$\pm$0.54
& 47.85\scriptsize$\pm$0.09
& 53.15\scriptsize$\pm$6.51
& 51.50\scriptsize$\pm$4.18
& \underline{72.57\scriptsize$\pm$3.31} 
& \textbf{77.88\scriptsize$\pm$2.81} \\
& 3$\rightarrow$2 
& 49.38\scriptsize$\pm$5.05 
& 74.61\scriptsize$\pm$10.3 
& \textbf{77.69\scriptsize$\pm$9.31}
& 77.53\scriptsize$\pm$11.2 
& 70.22\scriptsize$\pm$10.0 
& 75.69\scriptsize$\pm$16.3 
& \underline{77.62\scriptsize$\pm$3.21} 
& 76.98\scriptsize$\pm$5.36 
& 43.62\scriptsize$\pm$0.97 
& 50.63\scriptsize$\pm$4.50 \\                 
& 2$\rightarrow$1 
& 49.22\scriptsize$\pm$0.12 
& \textbf{53.23\scriptsize$\pm$4.67} 
& 46.84\scriptsize$\pm$4.00 
& 47.48\scriptsize$\pm$4.32 
& 48.28\scriptsize$\pm$6.67 
& 49.11\scriptsize$\pm$2.60 
& 44.67\scriptsize$\pm$3.12 
& 48.12\scriptsize$\pm$3.50 
& \underline{49.23\scriptsize$\pm$0.07} 
& 47.92\scriptsize$\pm$0.00  \\
\cmidrule{2-12}

\multirow{-6}{*}{Boiler} & AVG & 56.15 
& \underline{63.68}
& 62.24 
& 62.26 
& 60.70 
& 56.42 
& 62.45 
& 63.60 
& 60.54 
& \textbf{63.93} \\

\bottomrule

\end{NiceTabular}
}

\label{tbl:all_results_dev_risk}
\end{table*}

%% file: tables/selected_hyper.tex
\begin{table}[!bt]
\caption{Selected values of the hyperparameters in the adopted datasets. The first row indicates the range of hyperparameter search. The range of $T$ is specified with respect to $L$, however, we reported the selected timesteps values.}
\resizebox{\columnwidth}{!}{
\begin{NiceTabular}{@{}c|cccccc@{}}
\toprule
       & $\beta_1$ & $\beta_2$ & $\beta_3$ & $\beta_4$ & $\lambda$ & $T$ \\ \midrule

Range & [0.1, 1] & [0.001, 1] & [0.001, 1] & [0.001, 1] & [0.5, 1) & [0, 0.5L] \\
\midrule
SSC    &  0.96 &   0.1  & 0.05 &  0.1  &   0.79     &     150   \\
UCIHAR &  0.78 &   0.1  & 0.20 &  0.1  &   0.90     &     14    \\
HHAR   &  0.80 &   0.1  & 0.05 &  0.1  &   0.52     &     14    \\
WISDM  &  0.98 &   0.1  & 0.05 &  0.1  &   0.72     &     6     \\
Boiler &  0.92 &   0.1 & 0.15 &  0.1  &   0.88     &     10   \\ 
\bottomrule
\end{NiceTabular}
}
\label{tbl:selected_hpr_values}
\end{table}

%% file: Figures/diff_mixups.tex
\begin{figure*}[!tb] 
  \begin{subfigure}[b]{0.24\linewidth}
    \centering
    \includegraphics[width=\linewidth]{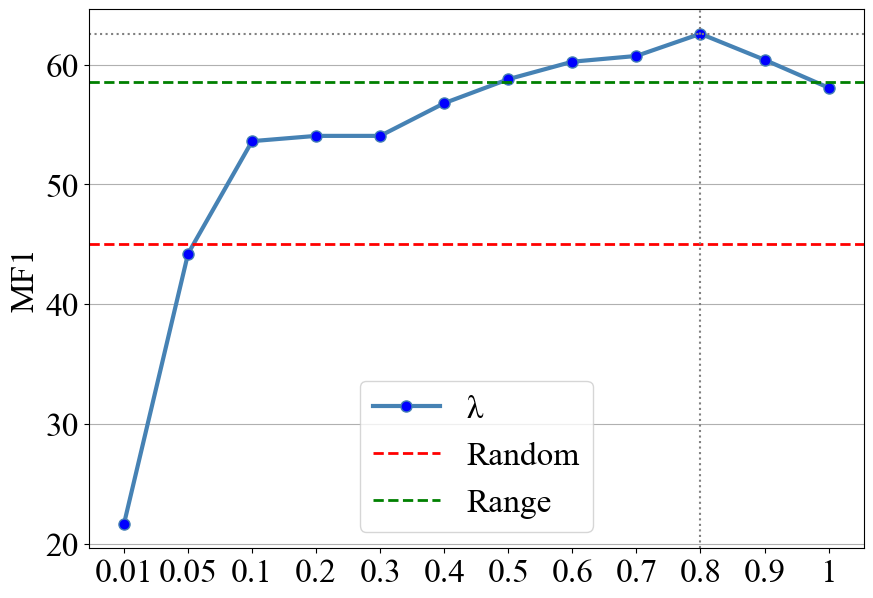}
    \caption{SSC}
    \label{fig:ssc_mixups} 
  \end{subfigure}
  \begin{subfigure}[b]{0.24\linewidth}
    \centering
    \includegraphics[width=\linewidth]{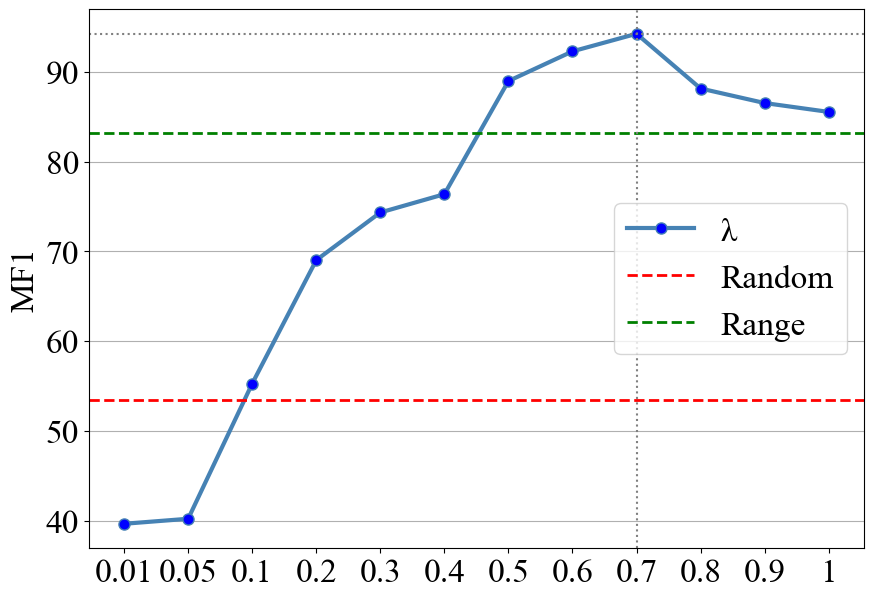}
    \caption{UCIHAR}
    \label{fig:har_mixups} 
  \end{subfigure} 
\begin{subfigure}[b]{0.24\linewidth}
    \centering
    \includegraphics[width=\linewidth]{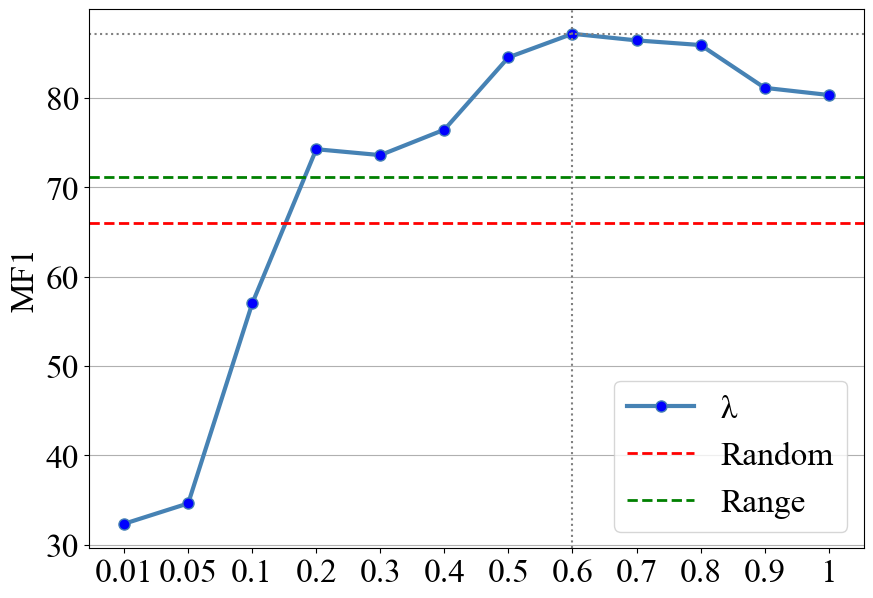}  
    \caption{HHAR} 
    \label{fig:hhar_mixups} 
  \end{subfigure} 
\begin{subfigure}[b]{0.24\linewidth}
    \centering
    \includegraphics[width=\linewidth]{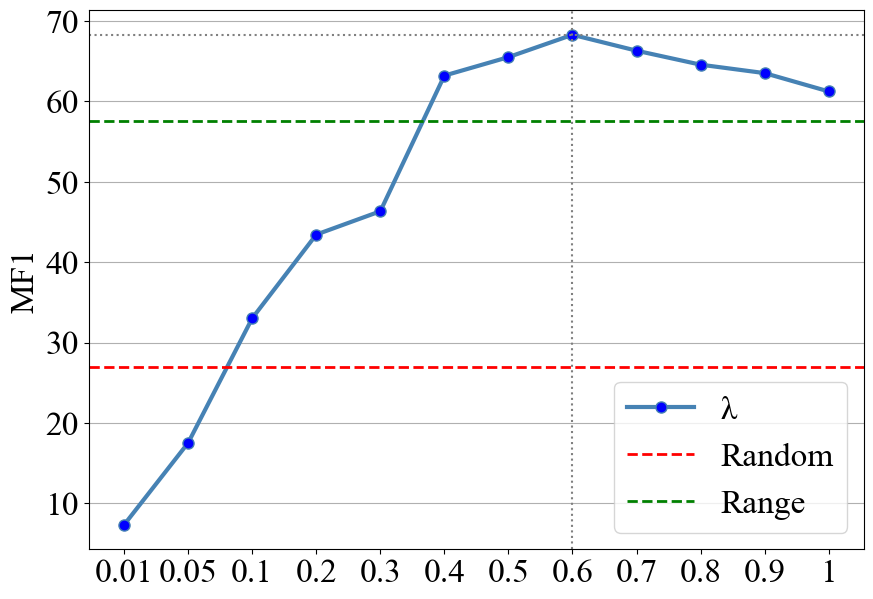}  
    \caption{WISDM} 
    \label{fig:wisdm_mixups} 
\end{subfigure} 
\caption{Study of different mixup strategies, as well as different values to our fixed temporal mixup strategy. The red dashed line indicates the average performance when selecting $\lambda$ randomly from a beta distribution. The green dashed line shows the average performance when selecting $\lambda$ randomly from a beta distribution but limited to a specific range $\geq 0.5$. The dashed lines show the $\lambda$ values achieving the best performance based on the target risk.}
\label{fig:diff_mixups}
\end{figure*}

%% file: tables/other_augmentations.tex
\begin{table}[!tb]
\centering
\caption{Comparison between deploying different augmentations against our temporal mixup for the in-domain contrastive adaptation. Clearly, other augmentations are not robust to the domain shift, and deploying them in contrastive adaptation yielded relatively less performance that our temporal mixup.}
\begin{NiceTabular}{@{}l|cccc@{}}
\toprule
\textbf{Augmentation} & \textbf{SSC} & \textbf{UCIHAR} & \textbf{HHAR} & \textbf{WISDM} \\ \midrule

Permutation & 57.15 & 80.63 & 76.79 & 54.67 \\
Scaling & 54.64 & 79.11 & 71.73 & 53.11 \\ 
Jittering & 56.53 & 80.27 & 75.23 & 51.04 \\
Masking & 56.09 & 80.44 & 74.64 & 55.34 \\
\midrule
\textit{Temporal Mixup} & \textbf{62.57} & \textbf{86.05} & \textbf{84.54} & \textbf{66.32} \\
\bottomrule
\end{NiceTabular}
\label{tbl:diff_aug_exp}
\end{table}

%% file: tables/risks_comparison.tex
\begin{table}[thb]
\centering
\caption{Comparison between the average performance with the realistic DEV risk and the overoptimistic target risk.}
\begin{NiceTabular}{@{}c|cc|cc@{}}
\toprule
 & \multicolumn{2}{c}{DEV risk} & \multicolumn{2}{c}{Target risk} \\ \midrule
Dataset & $\lambda$ & Avg. MF1 Score & $\lambda$ & Avg. MF1 Score \\ \midrule
SSC & 0.79 & 62.57 & 0.79 & 62.57 \\
UCIHAR & 0.90 & 86.05 & 0.70 & 94.26 \\
HHAR & 0.52 & 84.54 & 0.59 & 87.18 \\
WISDM & 0.72 & 66.32 & 0.60 & 68.30 \\ 
Boiler & 0.88 & 63.93  & 0.83 & 64.12 \\
\bottomrule
\end{NiceTabular}
\label{tbl:risks_comparison}
\end{table}

%% file: tables/ablation.tex
\begin{table}[!tb]
\centering
\caption{Ablation study showing the effect of each loss on the overall performance. By deploying contrastive loss on only one domain, it can still improve the performance. However, we clearly get the best performance by moving both domains towards an intermediate space.}
\resizebox{\columnwidth}{!}{
\begin{NiceTabular}{@{}ccc|cccc@{}}
\toprule
\multicolumn{3}{c|}{\textbf{Component}} & \multicolumn{4}{c}{\textbf{Dataset}}\\ \midrule

$\mathcal{L}_{\mathrm{ent}}$ & $\mathcal{L}_{\mathrm{CAC}}$ & $\mathcal{L}_{\mathrm{UC}}$ & SSC & UCIHAR & HHAR & WISDM  \\
\midrule

-  & - & - & 51.04 & 77.30 & 66.74 & 52.75 \\
\checkmark  & - & -  & 54.84 & 79.93 & 73.14 & 54.07  \\

\checkmark  & \checkmark  & -   & 59.14 & 83.82 & 77.34 & 58.79 \\
 
\checkmark  & -  & \checkmark &  57.43 & 80.87 & 80.11 & 61.11 \\

\checkmark  & \checkmark & \checkmark & \textbf{62.57} & \textbf{86.05} & \textbf{84.54}  & \textbf{66.32} \\

\bottomrule

\end{NiceTabular}

}

\label{tbl:ablation}

\end{table}

%% file: Figures/sens_figure.tex
\begin{figure}[!bt]
     \centering
     \begin{subfigure}[b]{0.49\columnwidth}
         \centering
         \includegraphics[width=\textwidth]{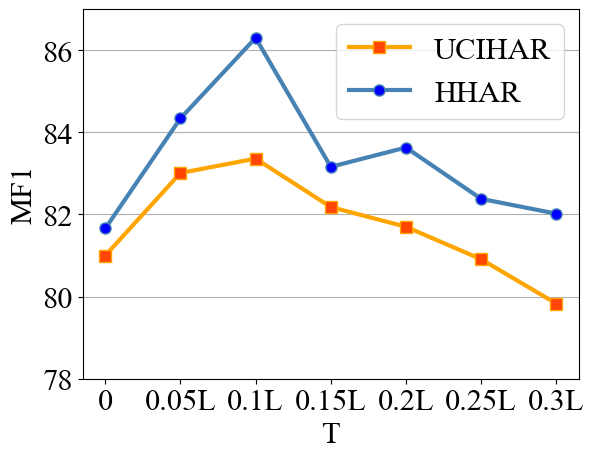}
         \caption{UCIHAR and HHHAR}
         \label{fig:ssc_sens}
     \end{subfigure}
     \hfill
     \begin{subfigure}[b]{0.49\columnwidth}
         \centering
         \includegraphics[width=\textwidth]{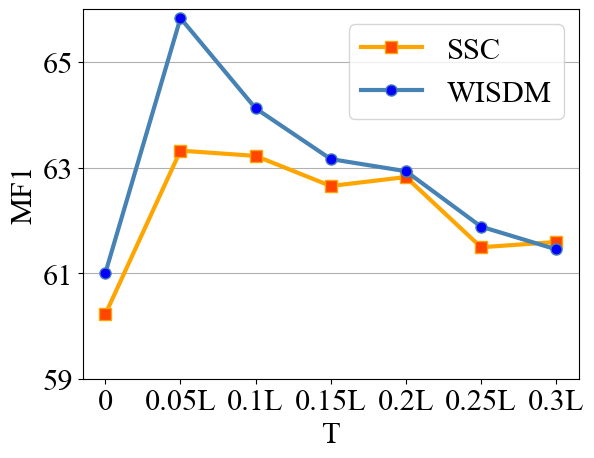}
         \caption{SSC and WISDM}
         \label{fig:har_sens}
     \end{subfigure}
     
\caption{Sensitivity analysis applied on the four datasets to study the effect of the number of timesteps $T$, as a percentage of the signal length $L$, on the performance. We notice a progressive performance improvement with including more timesteps in the temporal mixup operation until $T$ approaches $0.1L$. (We merged datasets with close performance in one figure)}
\label{Fig:sens_overall}
\end{figure}